\documentclass{amsart}
\usepackage{amsmath}
\usepackage{amsopn}
\usepackage{amssymb}
\usepackage{amsfonts}
\usepackage{graphicx}
\usepackage{boldline}
\usepackage{graphicx,booktabs}
\usepackage[table]{xcolor}
\usepackage{subfigure}
\usepackage{fancyhdr}
\usepackage[section]{placeins}
\usepackage [autostyle, english = american]{csquotes}
\usepackage{booktabs,dcolumn,caption}

\def\br{{\bf r}}
\def\bc{{\bf c}}
\def\bm{{\bf m}}

\usepackage[foot]{amsaddr}

\definecolor{maroon}{cmyk}{0,0.87,0.68,0.32}

\begin{document}

\title{\textbf{AUTOMATIC SHORT ANSWER GRADING AND FEEDBACK USING TEXT MINING METHODS} }
	
\author{N. S\"{u}zen}
\author{A.N. Gorban}
\author{J. Levesley}
\author{E.M. Mirkes}

\address{ Department of Mathematics, University of Leicester, Leicester LE1 7RH, UK }
\address{ Centre for Mathematical Modelling, University of Leicester, Leicester LE1 7RH, UK }

\email{ns433@le.ac.uk (N. Suzen)}
\email{ag153@le.ac.uk (A.N. Gorban)}
\email{jl1@le.ac.uk (J. Levesley)}
\email{em322@le.ac.uk (E.M. Mirkes)}

\maketitle

\begin{abstract}
Automatic grading is not a new approach but the need to adapt the latest technology to automatic grading has become very important. As the technology has rapidly became more powerful on scoring exams and essays, especially from the 1990s onwards, partially or wholly automated grading systems using computational methods have evolved and have become a major area of research. In particular, the demand of scoring of natural language responses has created a need for tools that can be applied to automatically grade these responses.

In this paper, we focus on the concept of automatic grading of short answer questions such as are typical in the UK GCSE system, and providing useful feedback on their answers to students. We present experimental results on a dataset provided from \textit{the introductory computer science class} in the University of North Texas. We first apply standard data mining techniques to the corpus of student answers for the purpose of measuring similarity between the student answers and the model answer. This is based on the number of common words. We  then evaluate the relation between these similarities and marks awarded by scorers. We consider an approach that groups student answers into clusters. Each cluster would be awarded the same mark, and the same feedback given to each answer in a cluster. In this manner, we demonstrate that clusters indicate the groups of students who are awarded the same or the similar scores. Words in each cluster are compared to show that clusters are constructed based on how many and which words of the model answer have been used. The main novelty in this paper is that we design a model to predict marks based on the similarities between the student answers and the model answer.

We argue that computational methods be used to enhance the reliability of human scoring, and not replace it. Humans are required to calibrate the system, and to deal with situations that are challenging. Computational methods can provide insight into which student answers will be found challenging and thus be a place human judgement is required.
\vspace{5mm}

\noindent {\textit{Keywords:}} automatic grading, machine learning, text mining, similarity measures, clustering, $ k $-means
\end{abstract}

\section{Introduction}

In the learning process, the assessment of knowledge plays a key role for effective teaching \cite{Mohler2}. With the range of assessment methods available, examinations have dominated the assessment of student learning. In particular, knowledge and understanding in academic courses are assessed by end of course examinations combined with coursework.  Academic examinations can be performed using many question types from multiple choice questions to free responses. Manual assessment is much more difficult for question types such as short answer and essay questions \cite{Reynolds}. Students are required to give \textit{free text responses} for these question types, so each response requires textual understanding and analysis as opposed to grading answers with a single correct answer such as multiple choice tests. Hence, manual scoring takes a considerable amount of time, and provision of meaningful feedback even more so.

Manual scoring of those answers can suffer from inconsistency since the marker must infer meaning from the candidates' own words. Scores on the same answer may vary from marker to marker. However, free text questions are a widely preferred assessment tool, used throughout the learning process, due to their effectiveness on developing cognitive skills of students and also demonstrating knowledge in short texts \cite{McDaniel}. Therefore, there is a need to develop tools for mitigating these challenges in assessment. One approach is to create automatic scoring tools and feedback mechanisms that supports markers. This methodology was discussed by several researchers, especially from the 1990s onwards, as computational techniques became applicable in this field. A number of studies have been made to automate short answer grading \cite{Leacock,Mitchell,Sukkarieh}.

\begin{figure}[b]
  \includegraphics[width=0.9\linewidth]{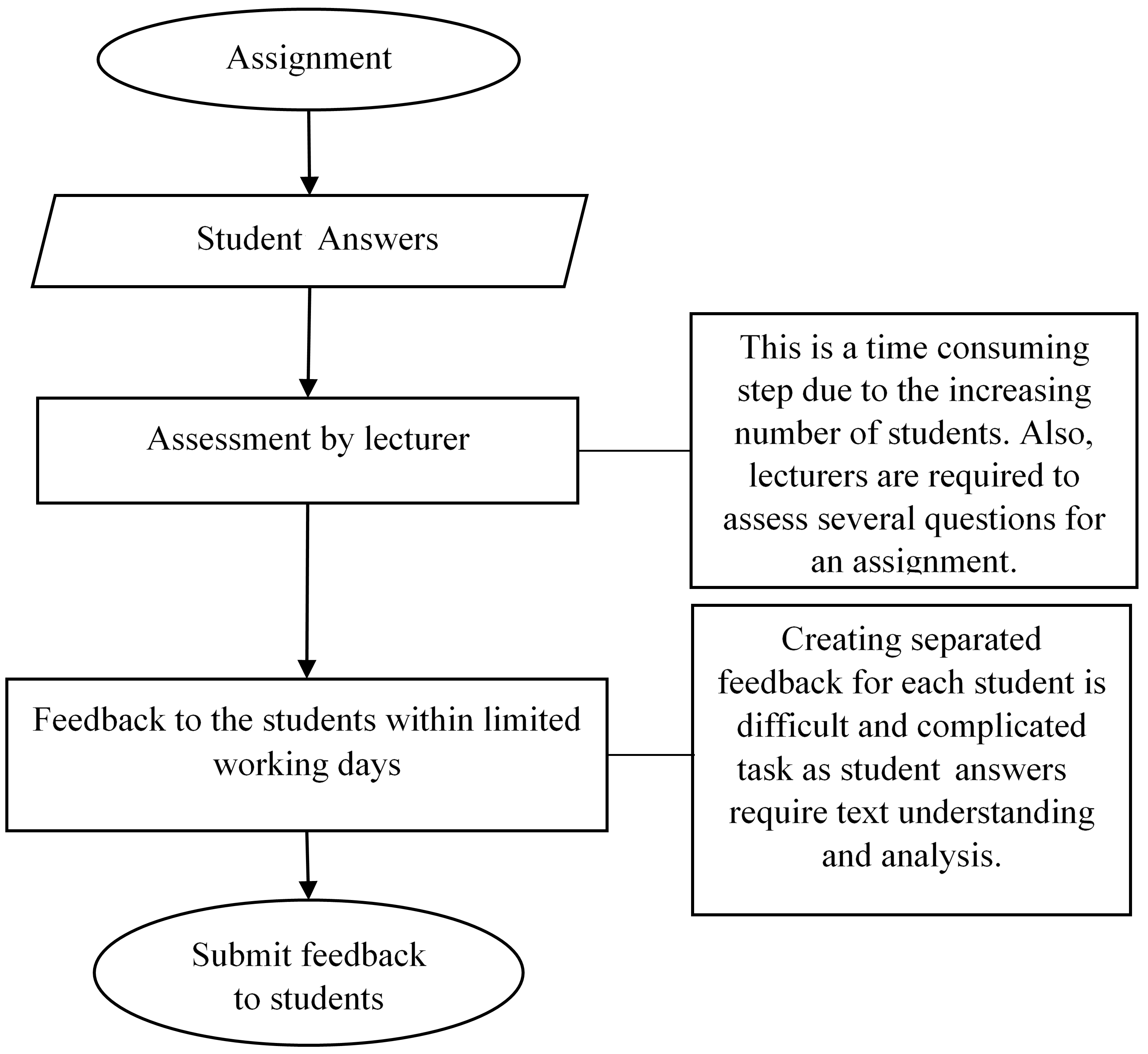}
  \caption{Steps and challenges in manual assessment}
  \label{fig:fig01}
\end{figure}

Assessment of natural language responses is a challenging task since we can not expect a machine to understand free text answers. However, developments in natural language processing (NLP) have made partially or wholly automated scoring of exams possible. Automatic grading has become a popular field among researchers due to its benefits on reducing human mistakes and time spent (see Figure~\ref{fig:fig01}).

Automatic assessment can also be considered for pre-assessment of student works by giving them support to improve their work. A student could submit their work to a feedback system and be given information such as: \textquotedblleft Answers like this scored this mark. In order to improve this you might ...". By tracking student use of such a system and we also have the possibility of tutor understanding of what is being misunderstood in class, with the opportunity for face to face mitigation.

This research presents both supervised and unsupervised approaches that deal with the automatic marking and feedback for short answers. The proposed models are based on the concept of similarity between the model answer and student answers, and the discovery of the structure in the corpus of student responses.  Our initial assumption is that given a set of student answers, marks awarded by scorers are highly dependent on words that the student used which also occur in the model answer. This is because in practice, grades are often awarded based on how similar a student response is to the expected answer (model answer). Using these similarities, we intend to build our model to mark student work.

We will see that this approach is more appropriate for some questions than others, but that questions that the algorithm finds hard to mark are also difficult for humans; the variation of marks of two human markers is higher.

The model calculates the \textit{distance} between the model answer and the student answer using words in the model answer  to generate scoring rules. This has the additional benefit of identifying the misconceptions and weaknesses of students for a topic under the assumption that those students who have a lack of knowledge are not able to use all of the words (or synonyms) in the model answer. Depending on the similarities between answers, automatic feedback can be given to make students aware of their level of understanding. Such a system can be also used as a supervised learning process to predict an answer's score and automate marking by using a training set (see  Appendix~\ref{figures} for details about such approaches).

In the learning process, providing feedback to student also plays a key role, as it helps students understand the subjects and improve their learnings and self-awareness. Our second approach focuses on an \textit{automatic feedback mechanism}. We cluster student answers into groups to explore whether or not responses given by students share similar characteristics. This approach can uncover natural groups of answers having similar structures that students frequently use. We find clusters of similar answers, and then to evaluate elements of these cluster using both human and computational means. Thus we will provide teachers with information about the common answers that students give, since students generally answer questions in similar ways.

The main advantage of this approach is that a new system can be built that teachers can give a common feedback to students belonging to the same cluster \cite{basu}. This can be done by selecting a prototype answer(s) from each cluster. Grouping allows teacher to provide feedback for the prototype answer, and this feedback can be assigned to the entire cluster at once. It could be also used in a feedback system in which students submit answers with historical feedback, and use this feedback to improve their answers for final   submission of assignments. Further feedback on student behaviour could be garnered from student use of such a system.

Grouping similar answers together can be implemented in the automatic marking of groups. This approach can significantly cut down on the time for manual marking, and improve consistency in marking and feedback. In particular, once groups of answers are identified, the system assigns a common score to whole group by using human marking. This grouping process will not be 100\% reliable and there may be answers which are more difficult to classify. In this case human intervention is necessary. It is not our desire to remove the human from the marking process, just to improve consistency and allow humans to apply judgement in the difficult cases.

Finally, we develop a model to predict marks by using distances between the model answer and the student answers. We hypothesise that marks can be predicted by this distance between student answers and the model answer as marks are highly correlated with this distance. The objective of this approach is to show how distances from model answer can be used to mark student answers. In the results section we will see that the model lies close to the average of the scores of the two markers.

In the following section, we start with a detailed historical analysis on the automatic assessment of short answers. We describe the data set used, and then report an experimental study using supervised and unsupervised learning. We conclude by describing our methodology and results, and future work.

\section{Related Work}

The earliest study of automated grading dates back to the work of Page \cite{Page}. In this article, Page introduced his ideas on computational methods for grading student essays, and also on prospective roles of computers for grading.  He experimented with automatic evaluation of student essays with 276 essays written by high school students. His idea is based on correlation between basic characteristics of the essays and grades assigned by four teachers \cite{Wresch}. For each essay, overall quality is evaluated by the adding of the ratings by these teachers. To calculate approximated values of these actual ratings by computer, accessible variables to a computer are identified by teachers.

The results show that the computer grades are not distinguishable from human grades. Page states that, in the future, the computer-based judge will be better correlated with each human judge than the other human judges are. He also observes that such successful results from computerized grading may lead to the possibility of automated grading systems for the evaluation of essays. Finally, it is important to stress here that he also outlined the idea of `giving feedback', suggesting that a computer print-out could suggest that the student correct identified misspellings, syntax mistakes, and the overuse of certain words.

Since his study, automated grading of free text responses has become a popular area with the focus on marking essays rather than short answer questions.  However, more significant studies have been done since the start of the 1990s, as computational techniques and software technology have become more powerful \cite{Landauer}. The most well-known essay assessment systems are Project Essay Grade (PEG), e-rater, and Intelligent Essay Assessor \cite{Shermis}. In this article we are concerned with automatically marking of short answer questions, and therefore we present some similar systems for short answer grading in detail, rather than essay marking.

C-rater is a scoring engine designed to grade short content-based answers \cite{Leacock,Burrows}. The goal of c-rater is to match teacher with students answers in terms of their concepts. It is modelled to identify paraphrases of model answers as correct answers. These paraphrases are built by normalising a variety of responses related to four primary sources of variation among sentences: syntactic variation, pronoun reference, morphological variation and synonyms, and also variation caused by spelling error. Once c-rater matches the concepts found in the student answer with those found in the model answer(s), it assigns scores based on the number of matched concepts. It is reported that c-rater reached 84\% agreement with human graders for the scoring of reading comprehension responses. In their own words, it is stated that \enquote{if the teacher uses the same question for several classes or over several semesters, then the advantages of the initial effort are worthwhile}.

Similar work to ours has been carried out on the grading of short answers in \cite{Mitchell,Jordan,bachman,pulman,cutrone,thomas}. These grade answers by based on the similarity between the model answer and the student answer. Related studies can be found in \cite{Mohler2,Mohler}.

The idea of grouping similar responses has been introduced, in parallel to our approach, in \cite{basu}. In this work the approach is referred as \textit{Powergrading} due to its amplification of human effort for scoring. It is designed to group (and subgroup) responses by clustering techniques in order to make scoring partially automated. The proposed approach is based on the idea of using both human and the machine ability to score, under the assumption that groups of similar answers can be quickly marked by a human by considering the whole group at once. They also aim to discover patterns of misunderstanding among students, then to give comprehensive feedback to student answers in the same cluster.

Similar work is described by \cite{horbach}, where the approach relies on the parallel assumption that similar grades can be assigned to groups of similar answers. They use a clustering algorithm to create groups of answers and then assign a single grade to the whole cluster. Specifically, they used 1,668 short answers to 21 questions, with sample solutions and grades assigned by teachers, from a listening comprehension task for German language learners.

In their study, the extraction of features has been done by word $n$-grams, character $n$-grams and keywords \cite{ngrams}.  The cosine similarity between feature vectors and the centroid of each cluster is calculated, and those items which are the most similar to the centroid (with highest similarity value) are grouped into the cluster. For labelling of items in clusters, three different methods are evaluated for selecting the optimum response to be labelled: random item selection from each cluster, selecting the closest item to the centroid of the cluster, and selection of one item belonging to the majority label of the cluster. When item selection methods are compared, the results show that selection of the closest item to the centroid leads to higher accuracy than random selection methods in general.

\section{The Data and Features}
The data set we use is from the \textit{introductory computer science class} in the University of North Texas\footnote{The dataset was downloaded from the archive hosted at \\ http://lit.csci.unt.edu/index.php/Downloads. Available at (accessed 19 December 2019)\\  https://github.com/dbbrandt/short\_answer\_granding\_capstone\_project/tree/master/data/source\_data.}. It consists of 29 student answers from ten assignments and two exams. For each question, there is one model answer in the data. The model answers are mostly one sentence, but some of them contain a single word sentence (see example in Table \ref{table:tab1}). In this paper we will give these questions our own numbers, which differ from those in the original study. We give further examples of questions from this study in Appendix~\ref{questions}. We show examples of two questions, 1 and 2, for which our scoring model is reliable, and two questions, 3 and 4, for which it is less reliable. This correlates with teacher difficulty in marking. Further study is needed into what sort of questions teachers can reliably score, and which are more challenging.

We define the \textit{model vocabulary (or vocabulary)} for a question as the collection of words in the model answer. Table \ref{table:tab2} shows an example of model vocabulary and student answers containing words from the vocabulary. The student answers are manually scored by two teachers. Grades are given between 0 and 5, 0 for incorrect answers, 5 for correct answers, and from 1 to 4 for partially correct answers. We consider each grade as an individual label. For our intended approach, we used these labels to evaluate how words and marks are correlated and to see which marks occur in each cluster. There are discrepancies in the grades from the two teachers, and we use both grades for creating our model.

Before starting the analysis of the responses, we initially applied preprocessing steps to remove irrelevant characters (e.g. numbers, punctuation). We also remove so-called {\it stop words} such as {\it the, and, it} from both model answer and the student answers in order to improve computational efficiency. Additionally, stemming is performed to combine different versions of words into a root. For instance, {\it eat, eating, eaten} are all the same word as far as our method is concerned. This preprocessing step significantly reduces the number of words we deal with.

Finally, when applying our clustering algorithm, we performed some user defined pre-processing steps to reduce the high dimension of 163 (163 individual words). As almost all students used words appeared in the question sentence, we treat these words as stop words, and removed them from student answers. In addition, we observed that some words appear in only one or two answers. These words have no discriminative power for clustering and leads to sparse vectors in the algorithm. Sparsity refers to vectors with 0 frequencies in most of their inputs. We also removed those words appears in less than 10\% of answers in the corpus. After all pre-processing steps, there are 20 individual words in the collection of student answers, that is, the dimension of the vector space has become 20. R codes for producing clusters and processing the students answers with instructions for usage of the code are available in \cite{ghnes}.

To show the results here, we choose the first question as it represents the average answer length, with one sentence, and also unique answers from each student. Throughout the study, all analyses have been done by using this question, the model answer of the question, and 29 student answers.

\begin{table}[tb]
\centering
\caption{Examples of questions and model answers with different lengths  in the data. }
 \label{table:tab1}
\renewcommand\arraystretch{1.5}
  \begin{tabular}{ | l |  p{8cm}|}

    \hlineB{2}
     \rowcolor{maroon!10}
    \textbf{Question 1} & What is the role of a prototype program in problem solving?  \\ \hline
    \textbf{Model Answer 1} & To simulate the behaviour of portions of the desired software product. \\   \hline
  \rowcolor{maroon!10}
    \textbf{Question 2} & What is the stack operation corresponding to the enqueue operation in queues?  \\ \hline
    \textbf{Model Answer 2} & Push \\ \hlineB{2}
  \end{tabular}  
 \end{table}

\begin{table}[tb]
\centering
\caption{Examples of  model vocabulary and student answers }
  \label{table:tab2}
  \begin{tabular}{|  l | p{8cm} | }
    \hline
    \textbf{Model Answer 1} &  To simulate the behaviour of portions of the desired   software product. \\ \hline
     \textbf{Model Vocabulary 1} &  simulate, behaviour, portion,  desire,  software, product. \\ \hline
    \textbf{Student Answer} & High risk problems are address in the prototype program   to make sure that the program is feasible.  A prototype may also be used to show a company that the \textbf{software} can be possibly programmed.    \\   \hline
  \textbf{Student Answer} & it \textbf{simulates} the \textbf{behavior} of \textbf{portions} of the \textbf{desired} \textbf{software} \textbf{product}    \\   \hline
  \end{tabular}  
\end{table}

\section{Data representation and $k$-means clustering } \label{kmeans}
A starting point for applying data mining tools to unstructured text data is to transform the text into an appropriate set of data \cite{dm1,dm2}. In other words, text representations of a collection should be converted into numeric vectors (feature vector form) to be able to apply statistical methods on the data.

In our study, we used the \textit{Bag of Words (BW)} model for representation of texts. In this model, each document (answer in our case) is a collection (bag) of words. The idea of the BW model is to extract unique words from the collection of documents, and to treat these words as individual features. Each document is represented as a vector of word frequencies. Since the frequency of a word increases as the number of appearances of a term increases, this shows us how important a word is for a document. This representation is called \textit{term frequency (TF)}, which represents the relevance of a term to the corresponding document.

In the vector space method, documents are points of a high dimensional space, where each dimension (each feature) corresponds to one word. Each element of a vector indicates the position of a document in a particular dimension. So, distance measures tell us how far two points are in the vector space, i.e., the distance between two documents. In our work on clustering, we perform one of widely applicable distance measure: Euclidean distance (the sum of squares distance).

Suppose we have $N$ features in our vector space and $\br_1=(r_{11},r_{12},\cdots,r_{1N})$ and $\br_2=(r_{21},r_{22},\cdots,r_{2N})$ are the representations of two documents in our vector space. Then the distance between these documents is
$$
d(\br_1,\br_2)=\sqrt{(r_{11}-r_{21})^2+(r_{12}-r_{22})^2+\cdots+(r_{1N}-r_{2N})^2}.
$$

Note that as longer answers have more words than short answers, the number of non-zero entitles of features and also frequencies may be more compared with shorter answers in the vector representation. This does not mean the longer answer is more relevant. To adjust the effect of length, term frequencies should be normalized. Note that before implementing the clustering algorithm, the data is normalized in order to convert the frequency of terms to a common scale which allows for comparison. Given an answer $\br$, the $L_{2}$-normalization is defined as

\begin{displaymath}
\hat \br = {\br \over d(\br,{\bf 0})},
\end{displaymath}
since the length of a vector is its distance from the origin. Given these normalised vectors, we run a classical clustering algorithm, \textit{k-means}, in order to cluster student answers.

The $k$-means algorithm is one of the most popular clustering methods. The theoretical and algorithmic aspects have been studied by many researchers \cite{Bock,Cox,Fisher, Steinbach}. The idea behind this method is that all points in a vector space are separated into $k$ clusters, and each cluster is represented by its centroid vector (the sum of the vectors divided by their number). After defining $k$ centroids, each document is assigned to a cluster by using the distance $d$ (for us the Euclidean distance). Then, the centroids are recalculated until we find an optimal set of clusters based on some criterion function \cite{Luo}. 

Suppose that we have, after the $i$th iteration a set of $k$ clusters, and the $j$th cluster has $N_j$ points in it $\br_{j,\ell}^i, \ell=1,2,\cdots,N_j^i$. The distortion function is defined as
\begin{displaymath}
E_i=\sum_{j=1}^{k}\sum_{\ell=1}^{N_j^i}d(\br_{j\ell},\bc_{j})^2,
\end{displaymath}
where
$$
\bc_{j} = {1 \over N_j^i} \sum_{\ell=1}^{N_j^i} \br_{j\ell},
$$
is the centroid of the $j$th cluster \cite{Chiang,Kaufman}.

The algorithm seeks a partition of data set by optimizing the error criterion. The steps of $k$-means are as follows \cite{Xu} :

\begin{itemize}
\item Begin with randomly $k$-partition and calculate centroids for each cluster (initial centroids).
\item 	Assign each data point to nearest cluster (nearest-neighbour rule) by calculating the distance to the nearest centroid.
\item 	Re-calculate the centroids based on the current partition.
	\item Repeat the second and the third steps until there is no change between two iterations, that is, until the algorithm has converged.
\end{itemize}
We expect that $E_i$ will decrease as the algorithm proceeds ($i$ increases) until it converges to some minimum value.

\section{The marking and feedback system}

\begin{figure}

	\includegraphics[width=0.67\linewidth]{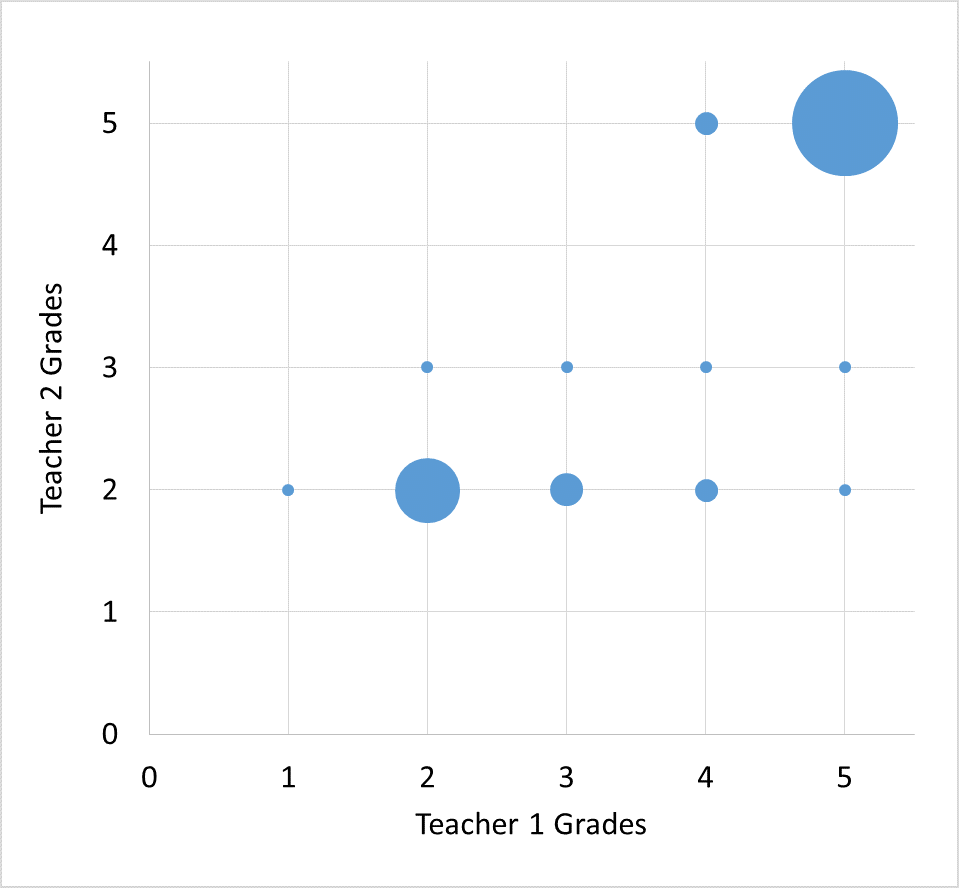}
	\caption{The distribution  of scores  awarded by two teachers. Size of points is proportional to number of cases, i.e, the proportion of pairs (teacher 1 grade, teacher 2 grade) in the dataset.}
	\label{fig:fig02}
\end{figure}

Our aim for this research is to design a model of automatic short answer marking and feedback. In this section, we will present an experimental study to demonstrate the processes behind the model creation. We begin with evaluating the dependency of scores and the similarity between the model answer and the student answer. Then, we will present our approach based on the clustering algorithm.

In the data used, not all of students are scored the same grades by the two graders. There are some cases where we observed inconsistency of grading.  Figure~\ref{fig:fig02} shows the distribution of scores awarded by the two teachers. The size of points indicates the proportion of grades for the corresponding pairs (Teacher 1 grade, Teacher 2 grade). As can be seen from the figure, there are 2 locations where the points are concentrated. The biggest proportion of scores comes from the pair (5,5) from both teachers, followed by the pair (2,2). This means that the teachers tended to agree with each other in grading for the most part (17 out of 29 students), especially when the answer is very good or very poor. Otherwise, there is inconsistency in scoring for some student answers. The correlation (Pearson correlation) between the two teachers' marks is 0.82 with an error of 0.1 (Mean Squared Error).

Note that the lowest and the highest grades given by teachers are 2 and 5 for this question, respectively. We suspect that those students graded at the lowest score did not use appropriate terminology for the subject. Similarly, those student graded highest answered used appropriate words. In between, they may use some of words required but the answer is not totally appropriate. We will now investigate why and how scores change depending on the words used.

\subsection{Unsupervised learning for clustering of student answers}

We now turn to looking at natural clusters of student answers to discover if there are some patterns of answers that students frequently use. If there are some groups in student answers, we can group them to give the same feedback or marks. Recall that we have treated words that appear in the question as stop words, and have removed these words from student answers (most of students include these words in their answers).

\begin{figure}
  \includegraphics[ width=0.7\linewidth]{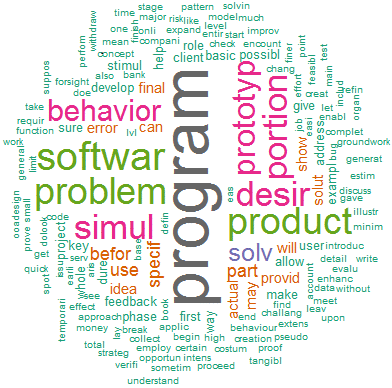}
  \caption{All words in the collection of student answers. The font and color of words indicate different frequencies of words.}
  \label{fig:fig5}
\end{figure}

In Figure \ref{fig:fig5}, all words that the students used in their answers can be seen with their frequencies. The more a word appears in the collection, the larger the font that is used in the word cloud; these are emphasised by the use of  different colors. For instance, the most frequently used word is \textquotedblleft program". This is an expected result since this word is contained in the question sentence; most of students started their answer with this word.

\begin{figure}
  \includegraphics[ width=0.95\linewidth]{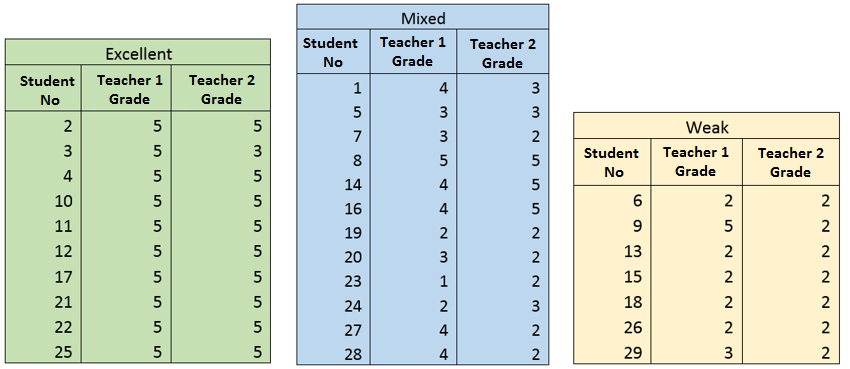}

  \caption{The structure of clusters with original marks awarded by two teachers.}
  \label{fig:fig11}
\end{figure}

To group answers, we use $k$-means clustering as described in Section~\ref{kmeans}. The optimal number of clusters is determined as three using the Elbow method \cite{park}. In Figure \ref{fig:fig11}, we show these three clusters with the associated grades awarded by the teachers. As can be seen from the tables, two clusters are well separated in terms of scores contained. We introduce clusters: {\it Excellent, Mixed} and {\it Weak} depending on the marks in the cluster. Student marks in the  cluster {\it Excellent} (shown as green) are 5, so we expect that these student answers are similar in terms of usage of the appropriate terminology. Similarly, answers in  cluster {\it Weak} (yellow) are expected to have inappropriate terminology since the marks are 2. However, we cannot identify a scoring rule for the  cluster \textit{Mixed }(blue), and further study based on keywords is required for this case. Note that there is also discrepancy in scores graded by two teachers in this cluster, so they are finding it challenging to score responses in this cluster. 

\graphicspath{ {SAM yazimis-upd/} }

\begin{figure}
  \includegraphics[width=0.95\linewidth]{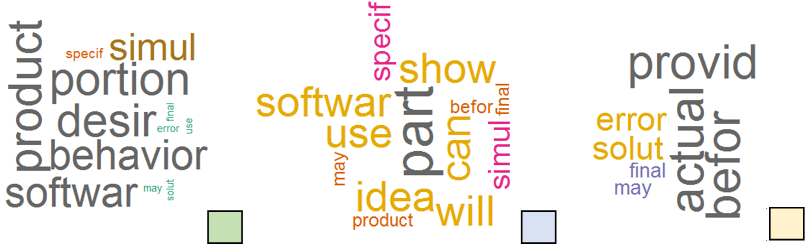}

\caption{Frequent words in each cluster. Colors in the bottom right hand corner show clusters (in the Figure \ref{fig:fig11}) that the words belong to.}\label{fig:test2}
\end{figure}
We now wish to examine what words are used by students in each group to understand the structure of groups. The frequently used words are displayed by word clouds; see  Figure \ref{fig:test2}. As expected, students in cluster {\it Excellent} used all of words from the model answer, while there does not exist any words from the model answer in the cluster {\it Weak}. When considering marks in these clusters, there are two marks: 5 and 2, respectively. This means that when students use all or none of the vocabulary words we are able to easily separate clusters for these students.

In addition, there are a few words from the vocabulary in the cluster mixed (software, simulate). We also see that some words such as `part' and `final' are synonymous with  `portion' and `desired' respectively. In this cluster, marks are spread with the highest value 5 and the lowest value 1 producing a range of 4. So, this shows that marks change depending words used in the cluster. We also note that a student has scored 5 from both teachers, but is not in our {\it Excellent} cluster. This shows that a better knowledge of acceptable vocabulary (synonyms) is needed in order to cluster more effectively. 

\subsection{Similarity between the model answer and the student answer} \label{simil}

Given that we can cluster responses and that the clusters correlate well with marks given, we wish to show how marks depend on the words used by students. We consider this problem as a basic calculation, where each answer is based on one model answer and two labels from two teachers. When we extract the significant words from the model answer we see that they number six (see Table~\ref{table:tab2}). 

We use the Hamming distance \cite{hamming} $h(\br,\bm)$ to measure the distance between student answer $\br$ and model answer $\bm$. We count the number of words $n$ that the students use which appear in the model answer. Then $h(\br,\bm)=6-n$, so that if all the words are used the distance is 0, and if none of the words are used, the distance is 6.

\begin{figure}
  \includegraphics[width=0.95\linewidth]{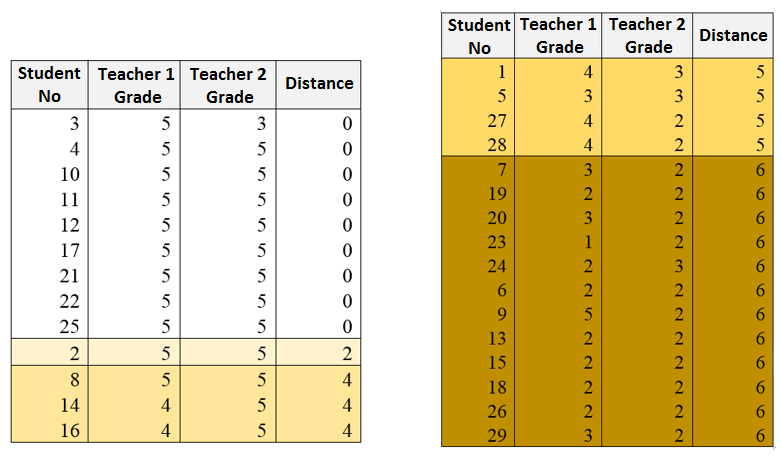}
  \caption{Distances between the model answer and student answers, and original marks for each student. Different colors indicate groups of students with different distances.}
  \label{fig:fig1}
\end{figure}

In a more sophisticated implementation we might need to look for synonyms of the words in the model answer, but for this study we are interested only in demonstrating the idea. The results are shown in the Figure \ref{fig:fig1}. We see that scores decrease for both graders in the main, in a regular way as the Hamming distance increases. It is also important to note that the scores are 5 when the distance is 0, and the majority of marks are 2 when the distance is 6. For other cases, we observe that the mark changes depending on the subset of words the student uses in their answer. Therefore, the level of importance of model words varies differently. Some of words may contribute more for marking. Teachers need to be able to set the importance of words for scoring for any automated system.

\graphicspath{ {SAM yazimis-upd/} }

We also computed the Pearson correlation coefficient to check the correlation between the distance and mark given, and found the correlation (Pearson correlation) -0.81 and -0.83 for two teachers with error 0.09 and 0.1, respectively. As expected, there is high correlation between marks given and the distances. In other words, we found that teacher assessment depends highly on how many of the vocabulary words students used in their answers.

\section{Supervised learning - a model to predict the mark}

In the previous sections we have demonstrated that the vocabulary used by students can be used to cluster their responses, and that the scores given by teachers are strongly related to the particular cluster that the student answer is assigned to. In this section we create and evaluate a model to predict student marks, based on the Hamming distance between the model answer and the student answer. We hypothesise that this distance is a strong indicator of the mark of a student, which suggests the possibility of automated scoring of responses. 

The relationship between the distance and the mark is modelled using the following predictor function

\[
  y= \beta_{0}+ \beta_{1} h^{ \beta_{2}  }, \tag{MM}\label{MM}
\]
where $ \beta_{0} $, $ \beta_{1} $, $ \beta_{2} $ are parameters, and $h$ is the distance between the model answer and the student answer. As shown in the Section~\ref{simil}, the distance can be 0, 2, 4, 5 or 6 for the question.

We fitted our model to the data by minimising the Mean Square Error of prediction (MSE) to find the optimal values of parameters \cite{curve1,curve2}. As there are inconsistencies in grading by the two teachers, we decided to use the average of their two grades as the actual value of the dependent variable. We call this teacher mark TM. Let us call this mathematical model MM to distinguish it from TM.

The estimates of parameters $ \beta_{0} $, $ \beta_{1} $, $ \beta_{2} $  and the MSE(MM) are presented in the Table \ref{turns}. The value of $ \beta_{0} $ represents the overall position of the curve on the $y$-axis; that is, the maximum mark which can be predicted by the model MM (when the distance is 0). We found a maximum mark of 4.91. The negative sign of  $ \beta_{1} $ tells us that scores decrease with distance. The value of $ \beta_{2} $  indicates how quickly the score decreases with distance from the the base line $ \beta_{0} $.

\begin{figure}[p]
  \includegraphics[width=0.7\linewidth]{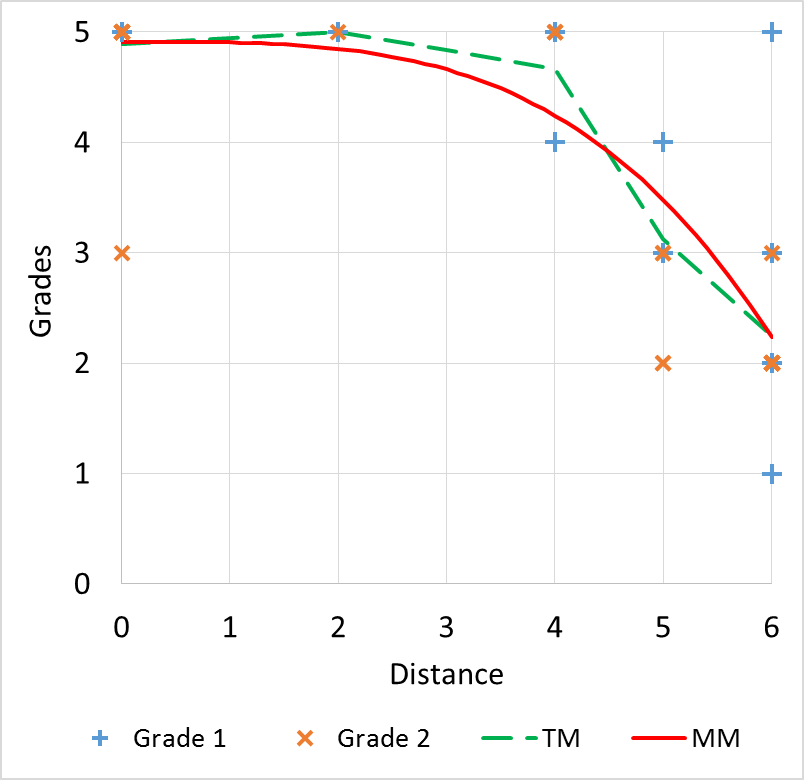}

  \caption{Distribution of actual marks of two teachers (Grade 1 and Grade 2) with distances from the model answer, the average marks of two teachers (TM) and the predicted marks by the mathematical model (MM).}
  \label{fig:figlast}
\end{figure}

 \begin{figure}[p]
  \includegraphics[width=0.7\linewidth]{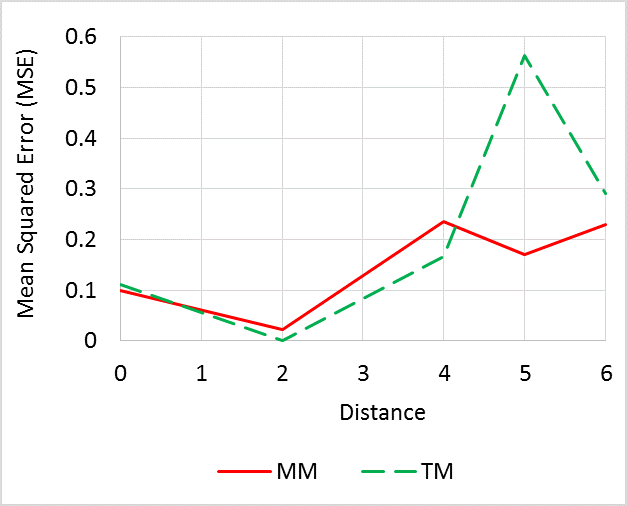}

  \caption{The mean squared errors for each distance from the model answer for MM and TM.}
  \label{fig:figlast2}
\end{figure}

\begin{table}[tb]

\centering
\caption{Estimation of parameters and the MSE for MM, $i=1,2,3,4$, for the corresponding questions (see Appendix~\ref{questions}); MSE is the mean square deviation of predicted marks from the average of two markers' grades. }

\begin{tabular}{|l|c|c|c|c|c|}
\hline 
Model & $ \beta_{0} $ & $ \beta_{1} $ & $ \beta_{2} $ & MSE (MM) & MSE (TM) \\ 
\hline
MM (1)      &  4.91085 & -0.0058 & 3.42359 & 0.17440 & 0.25000 \\ \hline
MM (2) &  5.00100 &	-0.00074 &	4.08469	& 0.37616	& 0.36290 \\  \hline
MM (3) & 5.03413	& -0.18210 &	1.92500	& 0.88490& 0.96667 \\ \hline
MM (4) & 4.54003	& -0.77408	& 0.40847	& 1.12573	& 0.43103 \\
\hline 
\end{tabular}
\label{turns}
\smallskip

\end{table}

To compare the accuracy of our model with human marking, we calculate the deviation of teacher grades from the average mark of two teachers. This approach is shown as TM (teacher marks) in the Table \ref{turns}. We see that the the teacher marks diverge from their average more than does the mathematical model for this question. This is unsurprising as it suggests the average is more predictable than the individual scores.

Figure \ref{fig:figlast} shows the observed data, the average marks of two teachers and the mathematical model versus distances from the model answer. In the figure, we see that for this question we have a line that mirrors the average of the teacher scores well. As expected, the mark is decreasing function of distance in general. Both models demonstrate qualitatively the same behaviour. To compare MM with TM model we graph the errors as a function of distance. These are depicted in Figure \ref{fig:figlast2}. We see that the accuracy of human marking depends on distance from the model answer. The disagreement between teachers increases significantly with the distance, until we regain agreement and accuracy for poor answers. In this  graph, we also look at the difference between these two predictions, and see it is small.

When the distance from the model answer is relatively small the MM perform well, with similar errors to the TM. On the other hand, MM outperforms TM model for big differences between the model answer and student answer (with distance 5 and 6).

Table \ref{turns} shows that the accuracy of the prediction of marks by MM is 0.17, and the estimation accuracy of TM is 0.25. This together with the Figure \ref{fig:figlast2} suggests that MM is no worse than human marking. 

We have performed a similar analysis on three other questions to demonstrate that some questions are harder to mark than others. 

\begin{figure}[p]
  \includegraphics[width=0.7\linewidth]{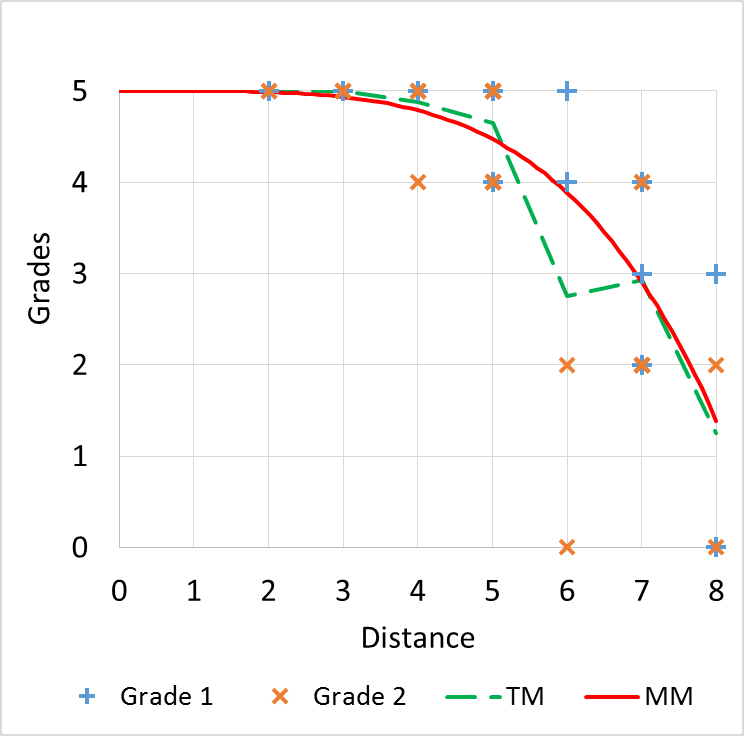}

  \caption{Distribution of actual marks of two teachers (Grade 1 and Grade 2) with distances from the model answer, the average marks of two teachers (TM) and the predicted marks by the mathematical model (MM) for the second question.}
  \label{fig:figlast17}
\end{figure}
\begin{figure}[p]
  \includegraphics[width=0.7\linewidth]{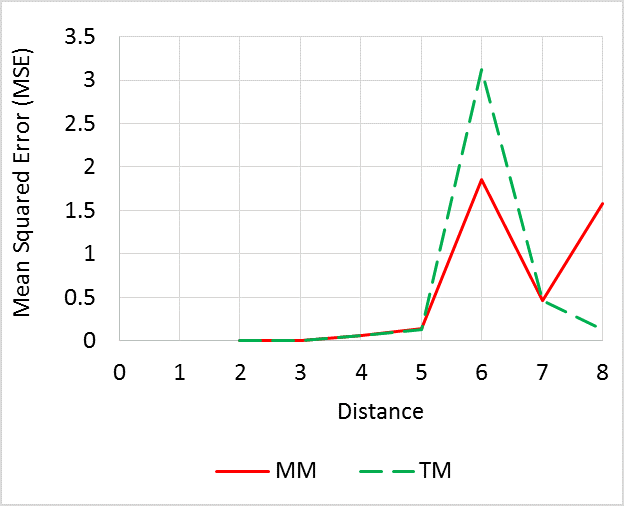}

  \caption{The mean squared errors for each distance from the model answer for MM and 	TM for the second question.}
  \label{fig:figlast172}
\end{figure}

We measure this by the extent to which the model mark deviates from the average of the teacher mark, and the extent to which teacher marks are a function of the distance of the answer from a standard word set. For Question 2, the corresponding results are shown in Figures~\ref{fig:figlast17} and \ref{fig:figlast172}, and show that for this question automated scoring agrees well with teacher scoring.

\begin{figure}[p]
  \includegraphics[width=0.6\linewidth]{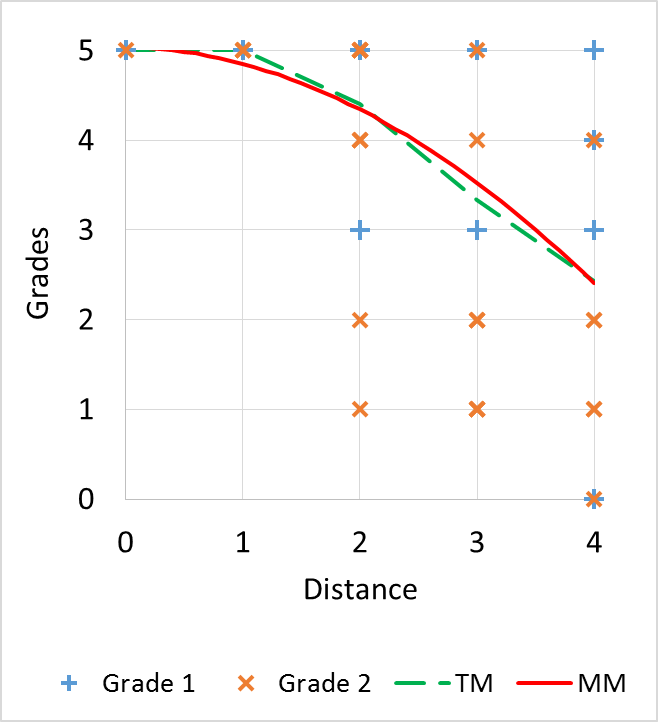}

  \caption{Distribution of actual marks of two teachers (Grade 1 and Grade 2) with distances from the model answer, the average marks of two teachers (TM) and the predicted marks by the mathematical model (MM) for the third question.}
  \label{fig:figlast26}
\end{figure}

Question 3 is an example which is harder to assess. In Figure \ref{fig:figlast26} we can see that our prediction is  close to TM with a slightly bigger error (MSE) for TM. However, the figure shows that marks vary from 1 to 5 for distances 2, 3 and 4. This contradicts to our  assumption that score is a function of distance. Therefore, disagreement of teachers on marking leads to inefficient scoring by the model MM. 

Another example is shown in Figure \ref{fig:figlast2_1}. We clearly see the absence of trend in original marks for this question (Question 4). The mark is not a decreasing function of distance. This contradicts to the assumption that as distance from the model answer increases, the mark given decreases. In this case, the model MM does not give reliable results. Another sign that MM cannot be effectively used for such examples is the value of  $\beta_{2}$. In MM, we expect $\beta_{2}>1$ and small $\vert  \beta_{1}  \vert$. With $\beta_{2} <1$ we have a change in the shape of the curve. 

 \begin{figure}[p]
  \includegraphics[width=0.7\linewidth]{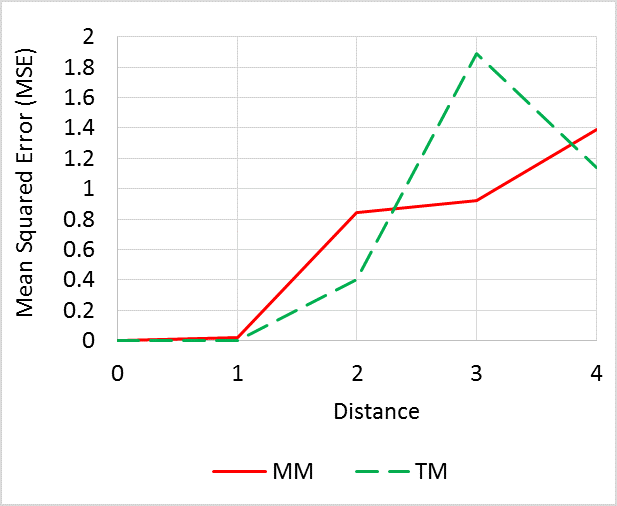}

  \caption{The mean squared errors for each distance from the model answer for MM and 	TM for the third question.}
  \label{fig:figlast262}
\end{figure}

\begin{figure}[p]
  \includegraphics[width=0.7\linewidth]{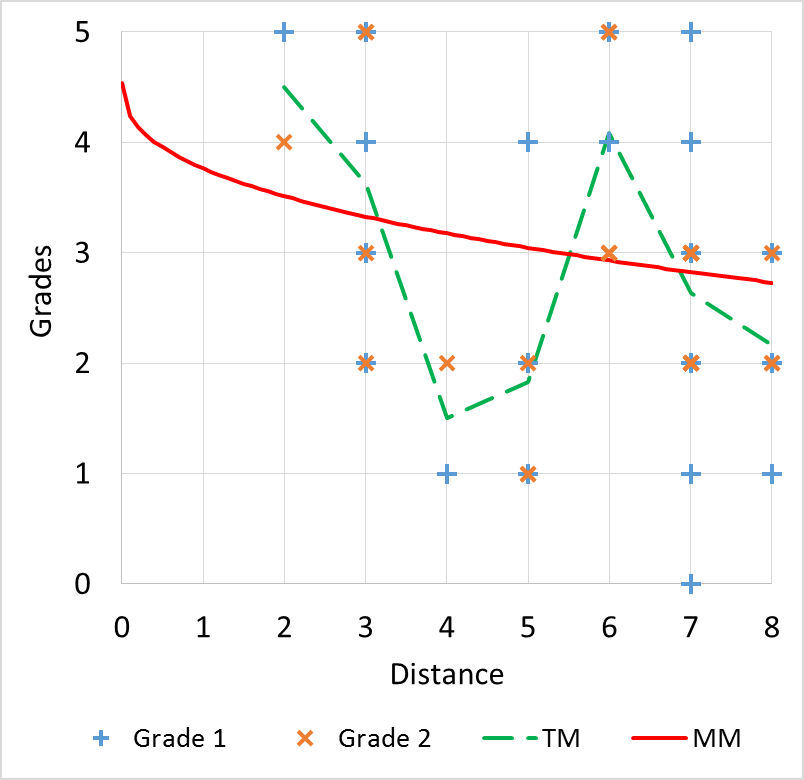}

  \caption{Distribution of actual marks of two teachers (Grade 1 and Grade 2) with distances from the model answer, the average marks of two teachers (TM) and the predicted marks by the mathematical model (MM) for the fourth question.}
  \label{fig:figlast2_1}
\end{figure}

 \begin{figure}[p]
  \includegraphics[width=0.7\linewidth]{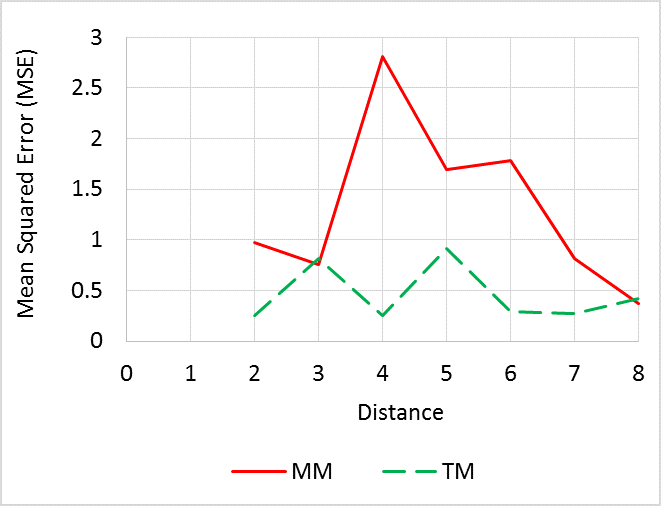}

  \caption{The mean squared errors for each distance from the model answer for MM and 	TM for the fourth question.}
  \label{fig:figlast2_2}
\end{figure}

\section{Conclusion and discussion}

This study sets out with the aim of developing a model for the automatic grading of short answer questions, and providing useful feedback to students. Experimental studies are presented to show how our approach succeeds with the automatic scoring of short natural language responses. Our methodology works well where a vocabulary for the model answer can be clearly identified. This can be automated in some cases, but may require human input.

Strong correlation of grades and Hamming distance from the model answer are found. We demonstrate correlation of 0.81 and 0.83 between this distance and grades of two human markers. The MSE of  linear regression for human marks are 0.09 and 0.1 for the two markers. This suggests that teacher marks can be predicted by distance with high accuracy. We conclude that the number of correctly used words has more influence on marks than semantics or order of words. In particular, if a large number of responses are being graded, it is not unreasonable that a human would move towards pattern recognition via key words rather than \textquotedblleft reading for meaning". Note that our system mirrors human scoring, and knows nothing about correct scoring. Hence, identifying words gives an idea about grades and students misunderstanding to teachers. Such an approach allows time saving for scoring, and to provide rapid feedback to students by checking the words used from model vocabulary.

We developed a model to predict marks using the distance between the model answer and the student answer. The proposed model has the form (MM), where parameters were estimated by minimising the mean squared error between the actual mark and the predicted mark. The average marks of two teachers are used  as observed marks since there is discrepancy of grades from the two teachers. The error of the prediction by the mathematical model (MM) is 0.17. The estimated accuracy of human grading (TM) is calculated as the mean deviation of actual marks from the average of the marks of the two graders, found as 0.25. MM has a lower deviation from the actual marks than TM.

We also consider a different approach for the automatic grading of short answer questions. This approach searches for natural groups of student answers. For each group, we can select one or more prototype mark (feedback), and assign it to the whole group at once. Clusters are found using $k$-means clustering. We note that we created clusters using word frequencies, without the use of teacher marks. 

The grades inside clusters indicate that our approach effectively identified groups containing the highest (5) and the lowest (1) grades. Analysis of the clusters shows that the first cluster contains only correct answers, the second contains of only incorrect answers, and the cluster with mixed grades contains all of the other answers. 

Finally, we tested our approach to see whether clusters are characterised by the frequency of words in the cluster and if this has any relation to the scoring rule. We show that there is a strong relationship between the clusters and the model vocabulary in students answers as well as grades. Those students who used all of the model words and those who used none of the model words, belong to different clusters. Grades inside clusters are similar. Such an approach provides teachers an opportunity to give common feedback and also grades to groups of students. The proposed approach also gives us the opportunity to look for common misconceptions in the answers given by students.

In order to improve our results, we need more input from the teachers, and  more detailed analysis. In our experiment, we observed spelling mistakes that should be corrected during the feedback process. We can detect again in the scoring process and adapt scores (if the teacher so wishes) related to spelling. We also observed that students used synonyms of the words in the model answer. A technical dictionary of synonyms could be developed (a standard thesaurus would contain too many unrelated words due to context), or teachers could provide acceptable alternative terminology.

Such an automatic scoring system can provide a clear baseline where conversations about assessment and feedback can develop. It is crucial that in this age of improving artificial intelligence, that we use machines to reduce the amount of repetitive straightforward scoring, which the human is poor at performing, and have people engaged in higher level, more valuable assessment and feedback.

\appendix
\section*{Appendices}
\addcontentsline{toc}{section}{Appendices}
\renewcommand{\thesubsection}{\Alph{subsection}}

\section{Questions from the University of North Texas Study} \label{questions}

The questions in this appendix are a sample of those from the \textit{introductory computer science class} in the University of North Texas \footnote{The dataset was downloaded from the archive hosted at \\ http://lit.csci.unt.edu/index.php/Downloads. Available at (accessed 19 December 2019)\\  https://github.com/dbbrandt/short\_answer\_granding\_capstone\_project/tree/master/data/source\_data.}. In Questions 1 and 2 we see that the teachers scores are more consistent and this means machine scoring such questions is more reliable. Questions 3 and 4 have more diverse answers and these are harder to score automatically.

\begin{table}[tb]
\centering
\caption{Example of a question for which we could reliably automate marking (Question 1).}
\renewcommand\arraystretch{1.5}
  \begin{tabular}{ | l |  p{8cm}|}

    \hlineB{2}
     \rowcolor{maroon!10}
    \textbf{Question 1} & What is the role of a prototype program in problem solving?  \\ \hline
    \textbf{Model Answer 1} &  To simulate the behaviour of portions of the desired   software product. \\ \hline
     \textbf{Model Vocabulary } &  simulate, behaviour, portion,  desire,  software, product. \\ \hline
    \textbf{Student Answer} & High risk problems are address in the prototype program   to make sure that the program is feasible.  A prototype may also be used to show a company that the \textbf{software} can be possibly programmed.    \\   \hline
  \textbf{Student Answer} & it \textbf{simulates} the \textbf{behavior} of \textbf{portions} of the \textbf{desired} \textbf{software} \textbf{product}    \\   \hlineB{2}
  \end{tabular}
   \label{table:tab11}
\end{table}

\begin{table}[tb]
\centering
\caption{Teacher marks for the answers to Question 1}
\renewcommand\arraystretch{1.5}
  \begin{tabular}{ | l | c | c | c | }

    \hlineB{2}
&	Teacher 1 grade &	Teacher 2 grade &	Average \\ \hline
1st student	& 4	& 3	& 3.5  \\ \hline
2nd student	& 5	& 5	& 5 \\ \hlineB{2}
\end{tabular}

\label{tab:tab11b}
\end{table}

\begin{table}[tb]
\centering
\caption{Example of a question for which we could reliably automate marking (Question 2)}
\renewcommand\arraystretch{1.5}
  \begin{tabular}{ | l |  p{8cm}|}

    \hlineB{2}
     \rowcolor{maroon!10}
    \textbf{Question 2} & How does the compiler handle inline functions?  \\ \hline
    \textbf{Model Answer 2} & It makes a copy of the function code in every place where a function call is made. \\   \hline
    \textbf{Model vocabulary} & Make, copy, function, code, every, place, call, made. \\ \hline
    \textbf{Student answer} & For inline functions, the compiler creates a {\bf copy} of the {\bf function's code} in {\bf place} so it doesn't have to {\bf make} a function {\bf call} and add to the function call stack. \\ \hline
    \textbf{Student answer} & It expands a small {\bf function} out... {\bf making} your {\bf code} longer, but also makes it run faster. \\ \hline 
    \textbf{Student answer} & The compiler can ignore the inline qualifier and typically does so for all but the smallest {\bf functions}. \\ 
   \hlineB{2}
  \end{tabular}
  
  \smallskip
    
  \label{table:tab4}
\end{table}

\begin{table}[tb]
\centering
\caption{Teacher marks for the answers to Question 2}

\renewcommand\arraystretch{1.5}
  \begin{tabular}{ | l | c | c | c | }

    \hlineB{2}
&	Teacher 1 grade &	Teacher 2 grade &	Average \\ \hline
1st student	& 5	& 5	& 5  \\ \hline
2nd student	& 4	& 4	& 4 \\ \hline
3rd student	& 2	& 2	& 2 \\ \hlineB{2}
\end{tabular}

\label{table:tab4b}
\end{table}

\begin{table}
\centering
  \caption{Question which is harder to assess (Question 3)}
\renewcommand\arraystretch{1.5}
  \begin{tabular}{ | l |  p{8cm}|}

    \hlineB{2}
     \rowcolor{maroon!10}
    \textbf{Question 3} & How many dimensions need to be specified when passing a multi-dimensional array as an argument to a function?  \\ \hline
    \textbf{Model Answer 3} & All the dimensions, except the first one. \\   \hline
    \textbf{Model vocabulary} & Dimension, except, first, one. \\ \hline
    \textbf{Student answer} & All {\bf dimensions} \textbf{except} for the {\bf first one} need to be specified when passing an array to a function, the compiler needs to know how many memory addresses to skip to make it back to the 2nd element in the first dimension.  The size of the first dimension does not need to be specified. \\ \hline
    \textbf{Student answer} & All {\bf  dimensions}, excluding the {\bf first one}. \\ \hline 
    \textbf{Student answer} & None, just pass the array name. \\ \hline
   \textbf{Student answer} & All of the {\bf dimensions} must be specified. \\ \hline
   \textbf{Student answer} & At least 2, depending on how many arrays are being used. \\
   \hlineB{2}
  \end{tabular}
  
  \smallskip
    
  \label{table:tab5}
\end{table}

\begin{table}[tb]
\centering
\caption{Teacher marks for the answers to Question 3}

\renewcommand\arraystretch{1.5}
  \begin{tabular}{ | l | c | c | c | }

    \hlineB{2}
&	Teacher 1 grade &	Teacher 2 grade &	Average \\ \hline
1st student	& 5	& 5	& 5  \\ \hline
2nd student	& 5	& 5	& 5 \\ \hline
3rd student	& 3	& 1	& 2 \\ \hline
4th Student	& 5	& 2	& 3.5 \\ \hline
5th Student	& 4	& 1	& 2.5 \\
\hlineB{2}
\end{tabular}

\label{table:tab5b}
\end{table}

\begin{table}[tb]
\centering
  \caption{Question which is harder to assess (Question 4)}
\renewcommand\arraystretch{1.5}
  \begin{tabular}{ | l |  p{8cm}|}

    \hlineB{2}
     \rowcolor{maroon!10}
    \textbf{Question 4} & What stages in the software life cycle are influenced by the testing stage?  \\ \hline
    \textbf{Model Answer 4} & The testing stage can influence both the coding stage (phase 5) and the solution refinement stage (phase 7). \\   \hline
    \textbf{Model vocabulary} & Test, stage, can, influence, code, phase, solute, refine.  \\ \hline
    \textbf{Student answer} & The implementation {\bf phase} and the maintenance phase are effected. \\ \hline
    \textbf{Student answer} & Elaboration, Construction, and Transition are all affected by {\bf testing}. \\ \hline 
    \textbf{Student answer} & {\bf Coding} and {\bf refining}. \\ \hline
   \textbf{Student answer} & 1- specification 2- design  3- risk analysis  4- verification  5- {\bf coding}  6- \textbf{testing} 7- {\bf refining}  8- production  9- maintenance. \\
   \hlineB{2}
  \end{tabular}
  
  \smallskip
  
  \label{table:tab3}
\end{table}

\begin{table}[tb]
\centering
\caption{Teacher marks for the answers to Question 4}
\renewcommand\arraystretch{1.5}
  \begin{tabular}{ | l | c | c | c | }

    \hlineB{2}
&	Teacher 1 grade &	Teacher 2 grade &	Average \\ \hline
1st student	& 5	& 3	& 4  \\ \hline
2nd student	& 2	& 2	& 2 \\ \hline
3rd student	& 5	& 5	& 5 \\ \hline
4th Student	& 4	& 1	& 2.5 \\ \hlineB{2}
\end{tabular}
\label{table:tab3b}
\end{table}

\pagebreak

\section{Feedback processes of the system} \label{processes}
\label{figures}
\FloatBarrier
\begin{figure}[!ht]
  \includegraphics[width=0.95\linewidth]{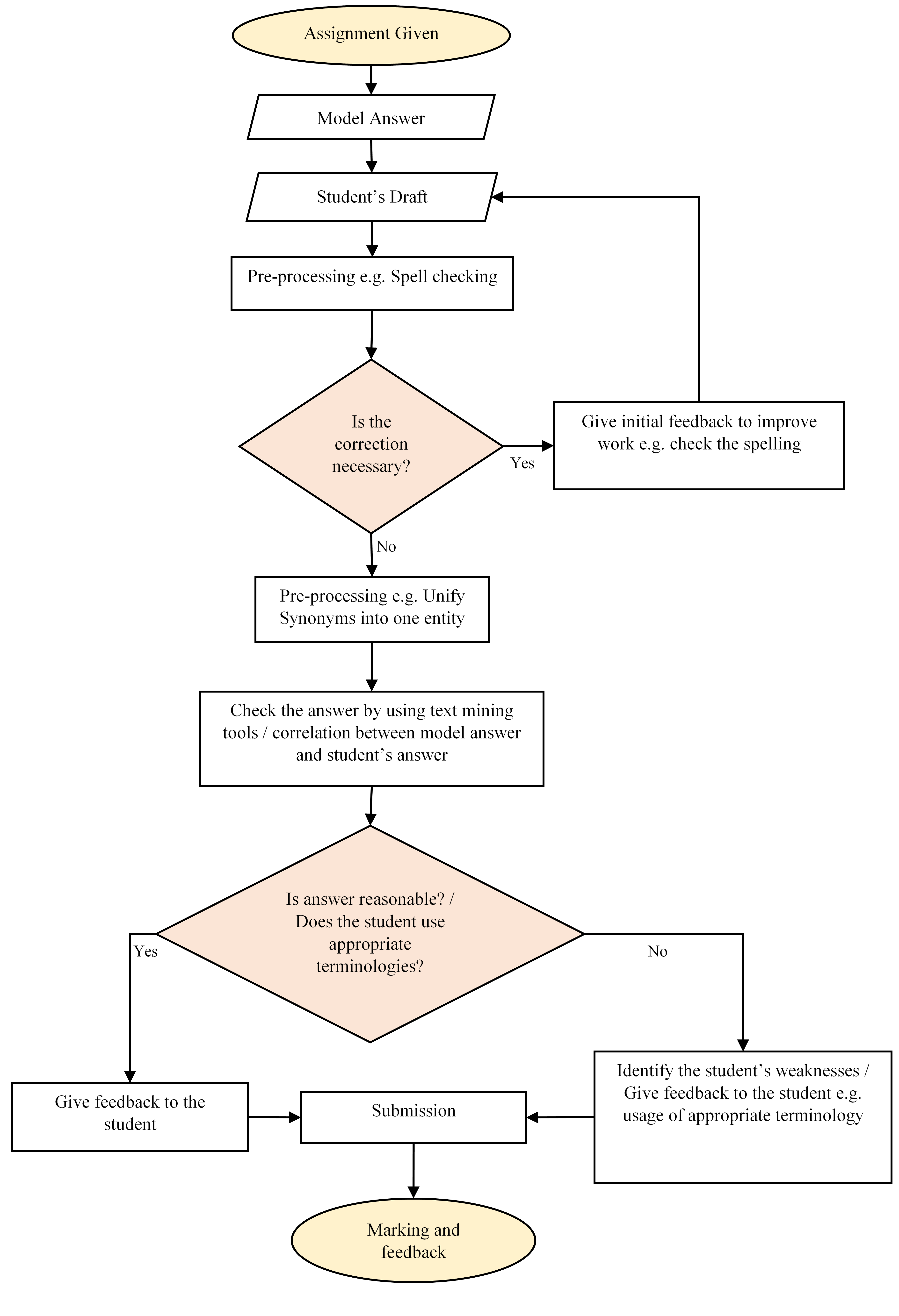}
  \caption{Feedback process by similarity between the model answer and students' answer}
  \label{fig:system1}
\end{figure}

	\begin{figure}
  \includegraphics[width=0.95\linewidth]{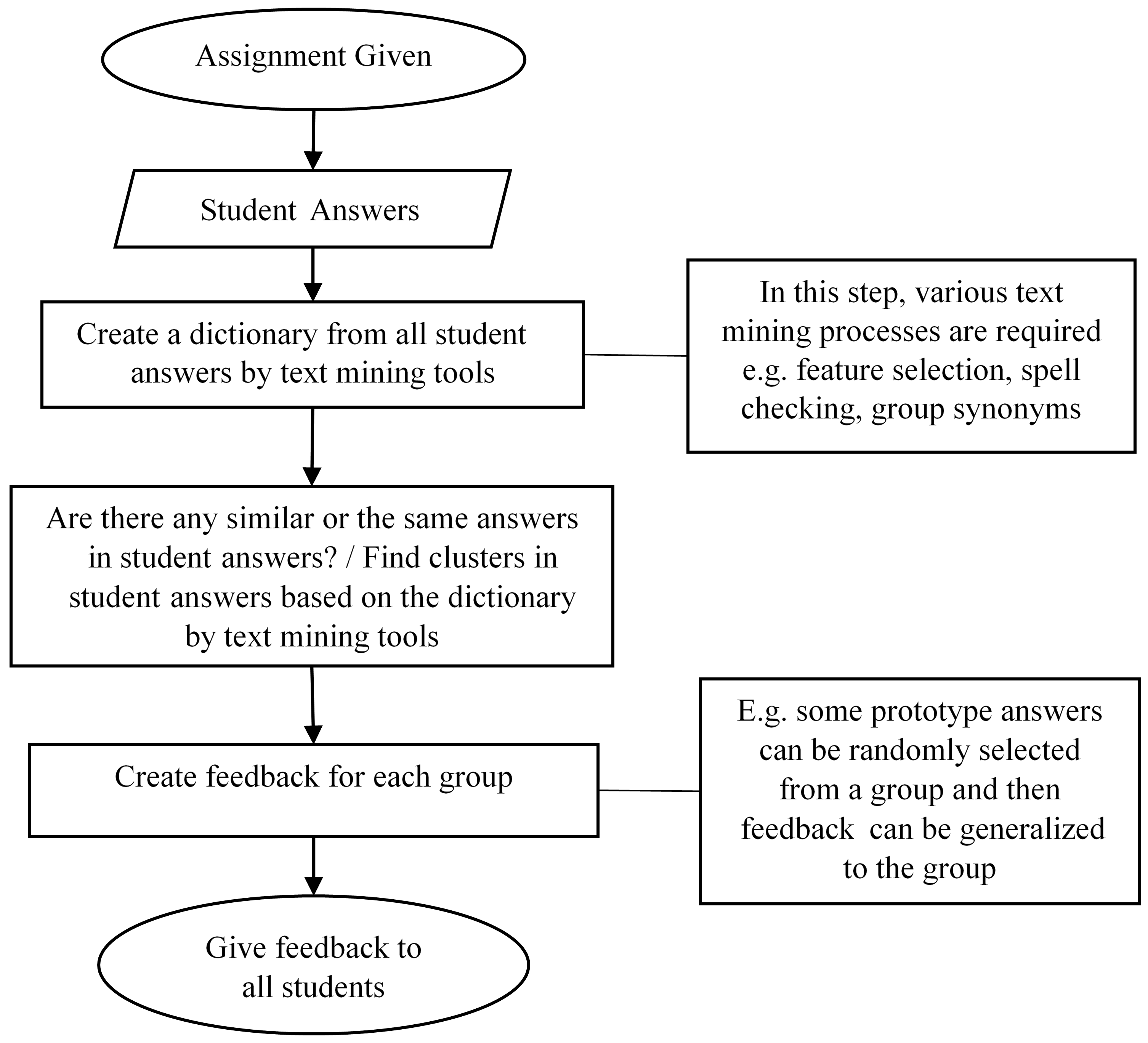}
  \caption{ Feedback process to group of students' answers by clustering approach  }
  \label{fig:system2}
\end{figure}
		
			\begin{figure}
  \includegraphics[width=0.95\linewidth]{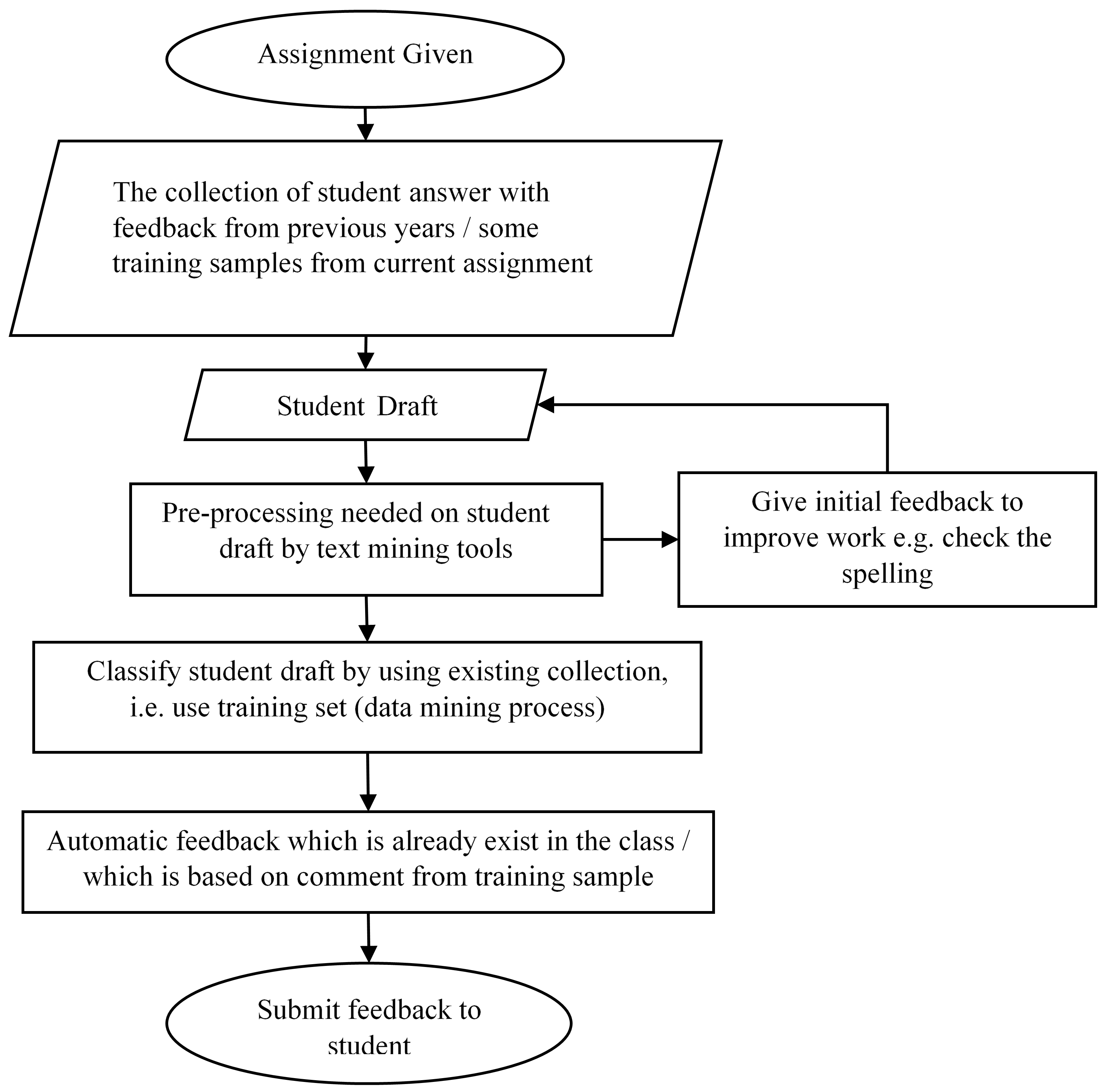}
  \caption{ Supervised process for feedback and marking in the case of existing train data   }
  \label{fig:system3}
\end{figure}
\end{document}